\documentclass[10pt,twocolumn,letterpaper]{article}

\usepackage{iccv}
\usepackage[accsupp]{axessibility}  % Improves PDF readability for those with disabilities.
\usepackage{times}
\usepackage{epsfig}
\usepackage{graphicx}
\usepackage{amsmath}
\usepackage{amssymb}
\usepackage{multirow}
\usepackage{booktabs}
\usepackage[font=normal]{subfig}
\usepackage{makecell}
\usepackage{bbm}
\usepackage{siunitx}
\usepackage{adjustbox}
\usepackage{tabu, booktabs}
\usepackage{makecell}
\usepackage{authblk}
\graphicspath{{figures/}}

\newcommand*{\affaddr}[1]{#1} % No op here. Customize it for different styles.
\newcommand*{\affmark}[1][*]{\textsuperscript{#1}}

\makeatletter
\newcommand{\printfnsymbol}[1]{\textsuperscript{\@fnsymbol{#1}}}
\makeatother

\usepackage[breaklinks=true,bookmarks=false]{hyperref}

\iccvfinalcopy % *** Uncomment this line for the final submission

\def\iccvPaperID{****} % *** Enter the ICCV Paper ID here

\ificcvfinal\fi

\begin{document}

%%%%%%%%% TITLE
\title{Preservational Learning Improves Self-supervised Medical Image Models by Reconstructing Diverse Contexts}

\author{%
	Hong-Yu Zhou\affmark[1]\thanks{First two authors contributed equally.}\quad Chixiang Lu\affmark[1]\printfnsymbol{1}\quad Sibei Yang\affmark[2]\quad Xiaoguang Han\affmark[4]\quad Yizhou Yu\affmark[1,3]\thanks{Corresponding author.} \\
	\affaddr{\affmark[1]The University of Hong Kong}\quad
	\affaddr{\affmark[2]ShanghaiTech University}\quad
	\affaddr{\affmark[3]Deepwise AI Lab}\\
	\affaddr{\affmark[4]Shenzhen Research Institute of Big Data, The Chinese University of Hong Kong (Shenzhen)}\\
	\tt\small{\{whuzhouhongyu, luchixiang\}@gmail.com, yangsb@shanghaitech.edu.cn, hanxiaoguang@cuhk.edu.cn, yizhouy@acm.org}\\
}

% \author{Hong-Yu Zhou\\
% The University of Hong Kong\\
% \and
% Chixiang Lu\\
% The University of Hong Kong\\

\maketitle
\ificcvfinal\thispagestyle{empty}\fi

%%%%%%%%% ABSTRACT
\begin{abstract}
Preserving maximal information is one of principles of designing self-supervised learning methodologies. To reach this goal, contrastive learning adopts an implicit way which is contrasting image pairs. However, we believe it is not fully optimal to simply use the contrastive estimation for preservation. Moreover, it is necessary and complemental to introduce an explicit solution to preserve more information. From this perspective, we introduce Preservational Learning to reconstruct diverse image contexts in order to preserve more information in learned representations. Together with the contrastive loss, we present Preservational Contrastive Representation Learning (PCRL) for learning self-supervised medical representations. PCRL provides very competitive results under the pretraining-finetuning protocol, outperforming both self-supervised and supervised counterparts in 5 classification/segmentation tasks substantially. Codes are available at \url{https://bit.ly/3rJydb1}.
\end{abstract}

%%%%%%%%% BODY TEXT
\section{Introduction}
It is common practice that training deep neural networks often requires a large amount of manually labeled data. This requirement is easy to satisfy in natural images as both the cost of labor and the difficulty of labeling can be acceptable. However, in medical image analysis, reliable medical annotations usually come from domain experts' diagnoses which are hard to access considering the scarcity of target disease, the protection of patient's privacy and the limited medical resources. To address these problems, self-supervised learning has been widely adopted as a practical way to learn medical image representations without manual annotations. 

\begin{figure}[t]
	\centering
	\includegraphics[width=0.7\columnwidth,height=0.33\columnwidth]{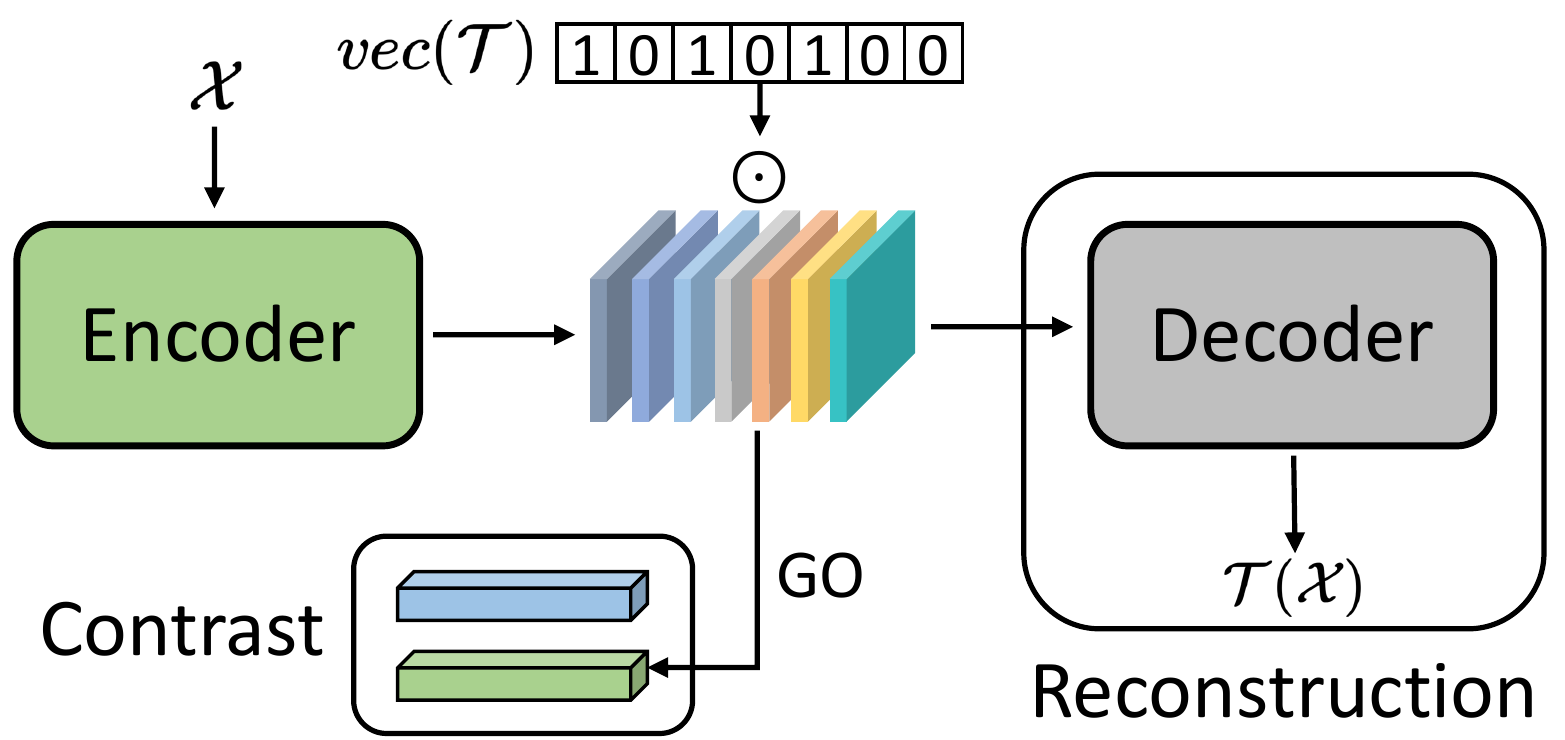}
	\caption{Conceptual illustration of proposed method. \textbf{GO} stands for global operations which convert feature maps to feature vectors. The blue feature vector comes from the momentum encoder. $vec(\mathcal{T})$ represents the indicator vector of $\mathcal{T}$ which contains a set of transformation functions. Each component in $vec(\mathcal{T})$ is 1 or 0 denoting whether the corresponding transformation is applied or not. $\odot$ stands for a channel-wise multiplication operation.}
	\label{intro}
\end{figure}

Nowadays, contrastive representation learning has been widely applied and outstandingly successful in medical image analysis \cite{zhou2020comparing,sowrirajan2020moco,chaitanya2020contrastive}. The goal of contrastive learning is to learn invariant representations via contrasting medical image pairs, which can be regarded as an implicit way to preserve maximal information. Nonetheless, we think it is still beneficial and complemental to explicitly preserve more information in addition to the contrastive loss. To achieve this goal, an intuitive solution is to reconstruct the original inputs using learned representations so that these representations can preserve the information closely related to the inputs. However, we discover that directly adding a plain reconstruction branch for restoring the original inputs would not significantly improve the learned representations. To address this problem, we introduce Preservational Contrastive Representation Learning to reconstruct diverse contexts using representations learned from the contrastive loss.

As shown in Fig.\ref{intro}, we attempt to incorporate the diverse image reconstruction, as a pretext task, into contrastive learning. The main motivation is to encode more information into the learned representations. Specifically, we introduce Transformation-conditioned Attention and Cross-model Mixup to enrich the information carried by representations. The first module embeds a transformation indicator vector ($vec(\mathcal{T})$ in Figure \ref{intro}) to high-level feature maps following an attentional mechanism. Based on the embedded vector, the network is required to dynamically reconstruct different image targets while the input is fixed. Cross-model Mixup is developed to generate a hybrid encoder by mixing the feature maps of the ordinary and the momentum encoders, where the hybrid encoder is asked to reconstruct mixed image targets. We show that both modules can help to encode more information and produce stronger representations compared to using contrastive learning only.

Besides the learning algorithm, this paper also addresses another issue when using unlabeled medical images for pretraining, that is lacking a fair and thorough comparison of different self-supervised learning methodologies. In this paper, we design extensive experiments to analyze the performance of different algorithms across different datasets and data modalities. Generally speaking, the contributions of this paper can be summarized into three aspects:
\begin{itemize}
	\item Preservational Contrastive Representation Learning is introduced to  encode more information into the representations learned from the contrastive loss by reconstructing diverse contexts.
	\item In order to restore diverse images, we propose two modules: Transformation-conditioned Attention and Cross-model Mixup to build a triple encoder, single decoder architecture for self-supervised learning.
	\item Extensive experiments and analyses show that the proposed PCRL has observable advantages in 5 classification/segmentation tasks, outperforming both self-supervised and supervised counterparts by substantial and significant margins.
	\end{itemize}

\section{Related Work}
In this section, we mainly review deep model based self-supervised learning approaches and mixup strategies. Note that for self-supervised learning, we only list the most related ones based on pretext tasks, ignoring clustering based approaches \cite{caron2018deep,zhao2020deep} and video based representation learning \cite{wang2015unsupervised,wang2017transitive,pathak2017learning,wang2019learning}.\\

\textbf{Pretext-based self-supervised learning in natural images.} Pretext-based methods rely on predicting input images' properties that are covariant to the transformations, such as recognizing image patches' content \cite{pathak2016context}, relative position \cite{doersch2015unsupervised,noroozi2016unsupervised}, rotation degree \cite{gidaris2018unsupervised,doersch2015unsupervised}, object color \cite{larsson2016learning, zhang2017split}, the number of objects \cite{noroozi2017representation} and the applied transformation function \cite{qi2020learning}. Contrastive-estimation based approaches also utilize pretext tasks to learn invariant representations by contrasting image pairs \cite{wu2018unsupervised,oord2018representation,chen2020simple,caron2020unsupervised,he2020momentum,zhou2017sunrise}. Recently, there are some works trying to remove the negative pairs in contrastive learning \cite{grill2020bootstrap,chen2020exploring}. By comparison, our method follows a different principle which is making representations able to fully describe their sources (\ie, corresponding input images). \\

\begin{figure*}[t]
	\centering
	\includegraphics[width=1.2\columnwidth, height=0.63\columnwidth]{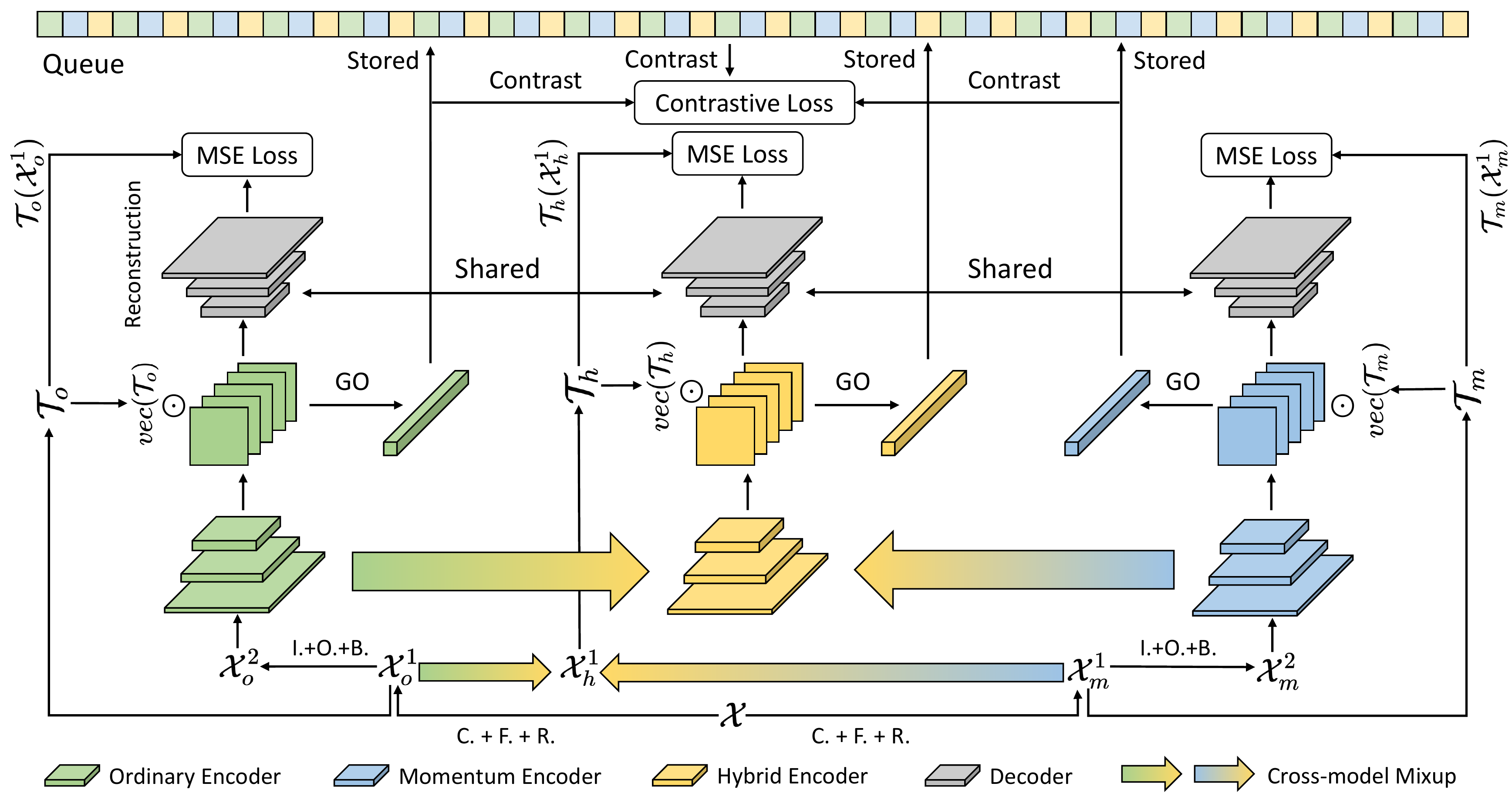}
	\caption{Overview of proposed framework. PCRL employs a U-Net like architecture to learn representations. For both encoder and decoder, we plot their feature maps for better demonstration. The hybrid encoder takes no input images as it consists of mixed feature maps from both the ordinary encoder and the momentum encoder. \{\textbf{C., F., R., I., O., B.}\} are abbreviations for random crop, random flip, random rotation, inpainting, outpainting and gaussian blur, respectively. \textbf{NCE} is short for noise-contrastive estimation. \textbf{GO} represents global operations which include global average pooling and fully-connected layers. $vec(\cdot)$ represents the indicator vector. \textbf{$\mathcal{T}_{\{o,m,h\}}(\cdot)$} denote a set of transformation functions for different encoders. $\odot$ represents channel-wise multiplication. For simplicity, we do not plot the skip connections.}
	\label{framework}
\end{figure*}

\textbf{Self-supervised learning in medical image analysis.} Before contrastive learning, solving the jigsaw problem \cite{zhuang2019self,zhu2020rubik,tao2020revisiting} and reconstructing corrupted images \cite{chen2019self,zhou2020models} are two major topics for pretext-based approaches in medical images. Besides them, Xie \etal \cite{xie2020nuclei} introduced a triplet loss for self-supervised learning in nuclei images. Haghighi \etal \cite{haghighi2020learning} improved \cite{zhou2020models} by appending a classification branch to classify the high-level features into different anatomical patterns. For contrastive learning, Zhou \etal \cite{zhou2020comparing} applied contrastive loss to 2D radiographs. Similar ideas have also appeared in few-shot \cite{zhou2021generalized} and semi-supervised learning \cite{zhou2021ssmd}. Taleb \etal \cite{taleb20203d} proposed 3D Contrastive Predictive Coding from utilizing 3D medical images. There are two works \cite{feng2019self,chakraborty2020g} most related to ours. Feng \etal \cite{feng2019self} showed that the process of reconstructing part images displays similar effects with those of employing a contrastive loss. Chakraborty \etal \cite{chakraborty2020g} introduced a denoising autoencoder to capture a latent space representation. However, both methods failed to improve contrastive learning with context reconstruction while our methodology succeeds in this aspect.\\

\textbf{Mixup in medical imaging.} Mixup \cite{zhang2017mixup}, as an augmentation strategy, has been widely adopted in medical imaging \cite{panfilov2019improving, chaitanya2019semi, jung2019prostate, eaton2018improving, bdair2020roam, wang2020focalmix}. The proposed Cross-model Mixup is most related to Manifold mixup \cite{verma2019manifold,jung2019prostate,bdair2020roam}. However, as far as we know, there is no previous method applying manifold mixup to cross-model representations, which is exactly the core contribution of our \textsc{Cross-model} Mixup.

\section{Methodology}
An overview of Preservational Contrastive Representation Learning (PCRL) is provided in Figure \ref{framework}. Generally, PCRL contains three different encoders and one shared decoder. The encoder and the decoder are connected via a U-Net like architecture. We first apply exponential moving average to the parameters of the \emph{ordinary encoder} to produce the \emph{momentum encoder}. Then, for each input, we apply Cross-model mixup to both encoders' representations (feature maps) to build a \emph{hybrid encoder}. Given a batch of images $\mathcal{X}$, we first apply random crop, random flip and random rotation to generate three batches of images $\mathcal{X}_o^1$, $\mathcal{X}_m^1$ and $\mathcal{X}_h^1$ for three different encoders, respectively. Then, we apply low-level processing operations, including inpainting, outpainting and gaussian blur, to each batch in order to generate the final inputs $\mathcal{X}_{\{o,m,h\}}^2$ for different encoders. In each training step, we randomly generate three sets of transformations (including flip and rotation): $\mathcal{T}_o$, $\mathcal{T}_m$ and $\mathcal{T}_h$ (please refer to Sec.\ref{sec31} for more details), and encode them into the last convolutional layer of each encoder. The ground truth targets of the MSE (mean square error) loss in image reconstruction are $\mathcal{T}_o(\mathcal{X}_o^1)$, $\mathcal{T}_m(\mathcal{X}_m^1)$ and $\mathcal{T}_h(\mathcal{X}_h^1)$, corresponding to different encoders. For contrastive learning in PCRL, we introduce noise-contrastive estimation which stores past representations in a queue \cite{he2020momentum} and then apply contrastive loss to both positive and negative image pairs.

\subsection{Transformation-conditioned Attention}
\label{sec31}
\begin{figure}[!htp]
	\centering
	\includegraphics[width=0.7\columnwidth]{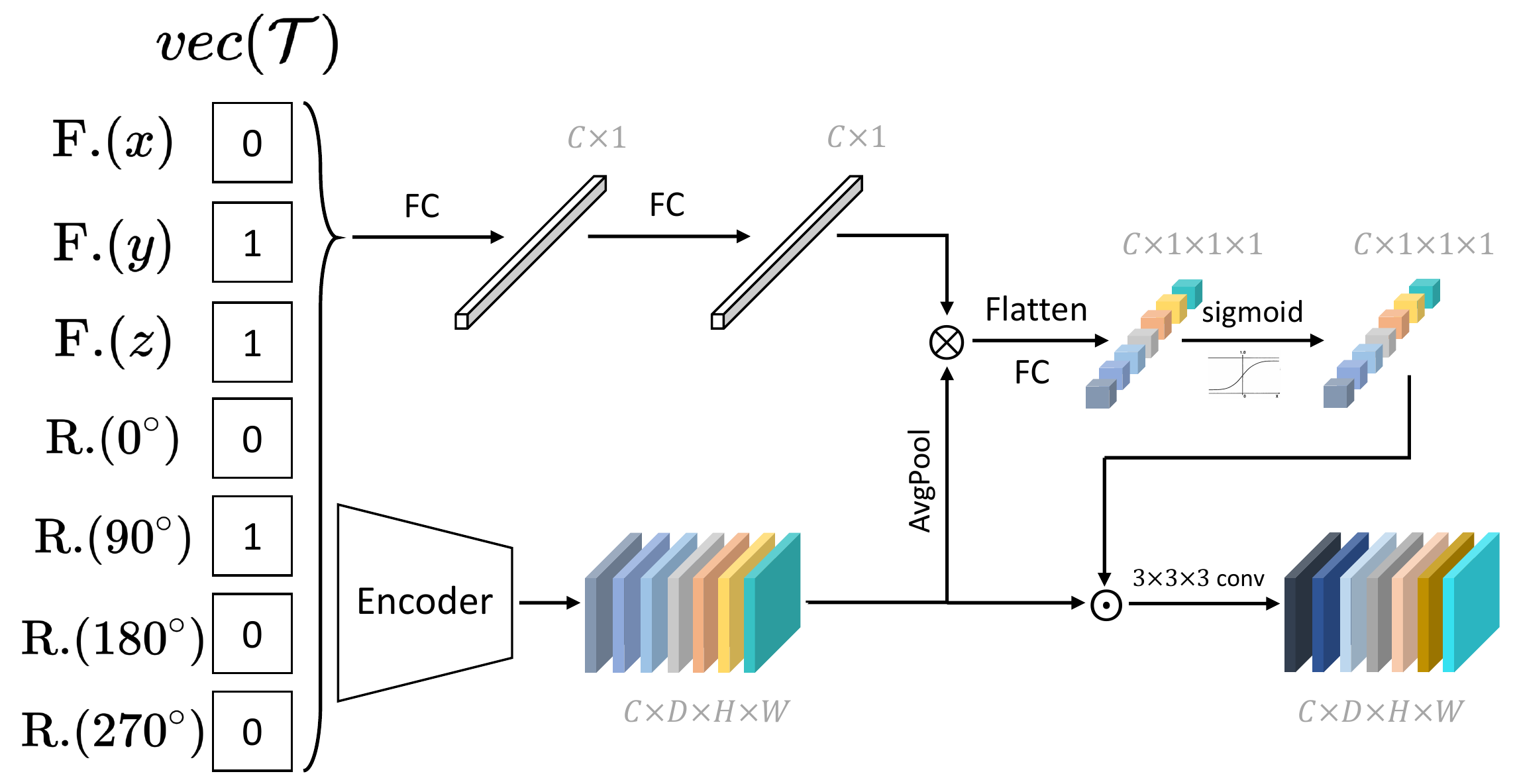}
	\caption{Our Transformation-conditioned Attention module. \textbf{F.} and \textbf{R.} stand for flip and rotation, respectively. \{$x,y,z$\} denote the axes. \{0,\ang{90},\ang{180},\ang{270}\} denote the rotation degree. $vec(\mathcal{T})$ denotes the indicator vector of $\mathcal{T}$ whose subscript is omitted for simplicity. $\otimes$ means the outer product. $\odot$ represents the channel-wise multiplication. Note that the above figure demonstrates the implementation when each input is 3D. For 2D inputs, there is no F.($z$) in the indicator vector. For both 2D and 3D inputs, the rotation is only applied to the xy-plane.}
	\label{TransAtt}
\end{figure}
In this section, we propose Transformation-conditioned Attention (TransAtt) to enable the reconstruction of diverse contexts. This module encodes the transformation vector into the high-level representations following an attentional mechanism. Such process can force the encoder to preserve more information in learned representations.

As shown in Figure \ref{TransAtt}, for each input, the indicator vector contains a combination of different transformations. Specifically, given 3D inputs (CT and MRI scans), the indicator vector has 7 components denoting different transformation strategies (cf. Figure \ref{TransAtt}). For 2D inputs (such as X-rays), the number of transformations decreases to 6 where F.($z$) does not exist. Each component contains an indicator function (1 or 0) representing whether the specific transformation is applied or not. To encode the indicator vector into high-level feature maps, we propose an attentional mechanism where we suppose that different channels of feature maps may have different impacts on the reconstructed results. Note that TransAtt is only applied to the last convolutional layer (before FC layers) of each encoder.

To imitate such process, we first forward the indicator vector $vec(\mathcal{T})$ to two fully-connected (FC) layers which produces a vector $f^p\in \mathbb{R}^{C\times 1}$. Meanwhile, we apply global average pooling to each encoder's high-level feature maps $\mathcal{F}^l\in \mathbb{R}^{C\times D\times H\times W}$\footnote{Here we omit the subscript $\{o,m,h\}$ which means $\mathcal{F}^l$ can represent feature maps from different encoders.} resulting in a vector $f^l\in \mathbb{R}^{C\times 1}$, where $l$ denotes the layer index. Then, we compute the outer product of $f^p$ and $f^l$:
\begin{align}
	\begin{split}
		M = f^p \otimes f^l,
	\end{split}
\end{align}
$M\in \mathbb{R}^{C\times C}$. Next, we flatten $M$ and forward it to another fully-connected layer:
\begin{align}
	\begin{split}
		f^q = \text{ReLU}\ (W_{\theta}\ \text{flat.}(M)),
	\end{split}
\end{align}
where $W_{\theta}\in \mathbb{R}^{C\times C^2}$ stands for the weight parameters of the FC layer. To perform rescaling, we further append a sigmoid function to $f^q$:
\begin{align}
	\begin{split}
		f^w=\text{sigmoid}(f^q),
	\end{split}
	\label{sigmoid}
\end{align}
where $f^w \in \mathbb{R}^{C\times 1\times 1\times 1}$. Finally, we apply channel-wise multiplication between $\mathcal{F}^l$ and $f^w$ and append a convolutional layer whose kernel size is 3:
\begin{align}
	\begin{split}
		\mathcal{F}^{l+1} = \text{conv}(\mathcal{F}^l \odot f^w),
	\end{split}
\end{align}
where $\mathcal{F}^{l+1}\in \mathbb{R}^{C\times D\times H\times W}$.

\subsection{Cross-model Mixup}
Apart from TransAtt, we introduce Cross-model Mixup (CrossMix) for shuffling these feature representations in order to enable more diverse restoration. Different from traditional mixup \cite{zhang2017mixup} which applies to network inputs, we propose to mix the feature maps from two different models to build a new hybrid encoder. 

Accordingly, the reconstruction target of the hybrid encoder is a mixed input $\mathcal{X}^1_{h}$. In practice, for each training iteration,
\begin{align}
	\begin{split}
		\mathcal{X}^1_h = \lambda \mathcal{X}^1_o + (1-\lambda)\mathcal{X}^1_m,
	\end{split}
	\label{alpha1}
\end{align}
where $\lambda\sim\text{Beta}(\alpha,\alpha)$, $\alpha$ is a hyperparameter\footnote{Beta distribution is employed in the original mixup paper.}. For network feature maps, we use $\mathcal{F}^{i}_{o}$ to denote the feature maps at layer $i$ of the ordinary encoder, $i\in \{1,...,l\}$. Similarly, $\mathcal{F}^{i}_{m}$ and $\mathcal{F}^{i}_{h}$ stand for the features maps at the same location in the momentum encoder and the hybrid encoder, respectively. Thus, the process of cross-model representation mixup can be formulated as:
\begin{align}
	\begin{split}
		\mathcal{F}^{i}_{h}=\lambda\mathcal{F}^{i}_{o}+(1-\lambda)\mathcal{F}^{i}_{m}.
	\end{split}
	\label{alpha2}
\end{align}
Together with the one shared decoder, we can directly use $\mathcal{F}_h^{\{1,...,l\}}$ to reconstruct $\mathcal{T}_{h}(\mathcal{X}_h^1)$.

\subsection{Loss Functions and Model Update}
To store past features for contrasting, we employ a queue $K=\{k_1,...,k_N\}$ to store them following \cite{he2020momentum}. The length of $K$ is $N$. In contrastive learning, we treat all features in queue $K$ as negative samples. Here we use $g_o(\cdot)$ and $g_m(\cdot)$ to denote the projectors of the ordinary encoder and the momentum encoder, separately. The contrastive loss $\mathcal{L}_c$ can be formulated as:
\begin{align}
	\begin{split}
		\mathcal{L}_c = -\text{log}\frac{\exp([g_o(\mathcal{F}^{l+1}_o)]^Tg_m(\mathcal{F}^{l+1}_m)/\tau)}{\sum_{j=1}^{N}\exp([g_o(\mathcal{F}^{l+1}_o)]^T k_j/\tau)},
	\end{split}
	\label{lc}
\end{align}
where $\tau$ is a temperature hyperparameter. $g_o(\cdot)$ and $g_m(\cdot)$ contains global average pooling and two FC layers, independently. After each training iteration, we push $g_o(\mathcal{F}^{l+1}_o)$, $g_m(\mathcal{F}^{l+1}_m)$, and $g_h(\mathcal{F}^{l+1}_h)$ to $K$ as negative samples for further contrasting. 

For reconstructing diverse contexts, we use mean square error (MSE) as the default reconstruction loss. Formally, if we denote the \emph{shared} decoder network as $D_{\theta}$, considering the whole network has a U-Net like architecture, the decoder's inputs should be multi-layer feature maps $\mathcal{F}_{\{o,m,h\}}$\footnote{We omit the superscript which is \{1,...,$l$+1\}.}. The computation of the reconstruction loss can be summarized as follows:
\begin{align}
\small
	\begin{split}
		\mathcal{L}_p &= \text{MSE}(D_{\theta}(\mathcal{F}_{o}),\ \mathcal{T}_{o}(\mathcal{X}_{o}^1)) + \text{MSE}(D_{\theta}(\mathcal{F}_{m}),\ \mathcal{T}_{m}(\mathcal{X}_{m}^1))\\
		&\ \ \ \ \ + \text{MSE}(D_{\theta}(\mathcal{F}_{h}),\ \mathcal{T}_{h}(\mathcal{X}_{h}^1)).
	\end{split}
	\label{ploss}
\end{align}
We finally sum up $\mathcal{L}_c$ and $\mathcal{L}_p$ as the complete loss function with equal weight (0.5 to 0.5). For network parameters, we denote the parameters of the ordinary encoder and the momentum encoder as $\theta_{o}$ and $\theta_{m}$, respectively. We update $\theta_{m}$ by using an exponential moving average (EMA) factor $\beta$:
\begin{align}
	\begin{split}
		\theta_m = \beta \theta_m + (1-\beta) \theta_o.
	\end{split}
\end{align}
Note that the hybrid encoder has no encoder parameters as it directly takes a combination of the feature maps from the ordinary and the momentum encoders. The mixed feature maps are then treated as the inputs to the shared decoder as shown in Equation \ref{ploss}.

\section{Experiments}
In this section, we first make ablation studies to demonstrate the advantages of TransAtt and CrossMix. Then, we introduce a thorough analysis of different self-supervised algorithms from different aspects. For all tasks, we employ the notation of \emph{source dataset}$\rightarrow$\emph{target dataset}. The source dataset is used for self-supervised pretraining while the target dataset is used for supervised finetuning. 

%More experiments (e.g., using more unlabeled data and larger models) and visual analysis can be found in the supplementary material.
\subsection{Baselines}
For medical pretraining approaches, we divide them into two categories: 2D and 3D, simply based on their input dimension (e.g., X-ray is 2D while CT scan is 3D). For 2D image pretraining, our baselines include train from scratch (TS), ImageNet pretraining (IN), Model Genesis (MG) \cite{zhou2020models}, Semantic Genesis (SG) \cite{haghighi2020learning} and Comparing to Learn (C2L) \cite{zhou2020comparing}. Here we ignore the method proposed in \cite{xie2020nuclei} which made prior assumptions on the number of nuclei and is not be suitable for other datasets. For 3D volume pretraining, we also include train from scratch (TS), Model Genesis (MG), Semantic Genesis (SG) and 3D-CPC \cite{taleb20203d}. Additionally, Cube++ \cite{tao2020revisiting} is included as it is an improved version of Rubik's Cube \cite{zhuang2019self} and Rubik's Cube+ \cite{zhu2020rubik}.

\subsection{Datasets}
In 2D tasks, we make experiments on two X-ray datasets: Chest14 \cite{wang2017chestx} and CheXpert \cite{irvin2019chexpert}. We use Chest14 for both 2D pretraining and 2D finetuning while CheXpert is only used for pretraining considering CheXpert contains a number of uncertain labels. The evaluation metric in 2D tasks is AUC. In order to evaluate the performance of algorithms on 3D volumes, we make experiments on CT and MRI datasets, including LUNA \cite{setio2017validation}, BraTS \cite{bakas2018identifying} and LiTS \cite{bilic2019liver}. We use LUNA for both 3D pretraining and 3D finetuning. The evaluation metric of finetuning on LUNA is AUC. BraTS is only used for supervised finetuning to test the cross-modal transferability following \cite{zhou2020models}. LiTS is mainly used for 3D finetuning on liver segmentation. The evaluation metric for segmentation is mean dice score. In practice, we divide each dataset into the training set, the validation set and the test set. The pretraining data always come from the training set (without labels). Please refer to the supplementary material for more details.

\subsection{Implementation Details}
We use 2D U-Net \cite{ronneberger2015u} and 3D U-Net \cite{cciccek20163d} as the backbone networks for 2D and 3D tasks, where we replace the encoder in 2D U-Net with ResNet-18. The EMA factor $\beta$ of updating momentum encoder is set to 0.99. For self-supervised pretraining, we employ momentum SGD as the default optimizer whose initial learning rate is set to 1e-3 while the momentum value is set to 0.9. We employ the cosine annealing strategy for decreasing learning rate and stop the training when the validation loss does not change for 30 epochs. The checkpoints with lowest validation loss values are saved for finetuning. For supervised finetuning, we use Adam as the optimizer with 1e-4 as the initial learning rate. Similar to pretraining, we rely on validation loss to determine when to end the training stage, and we save the checkpoints with lowest validation loss values for testing. Dice loss is used for segmentation tasks while cross entropy is employed for classification tasks. For other hyperparameters in baselines, we simply follow the choices in their official papers. $\alpha$ is set to 1 (for $\lambda$) in both Equation \ref{alpha1} and \ref{alpha2}. We set the temperature factor $\tau$ of softmax function in Equation \ref{lc} to 0.2 in practice. For each experiment, we repeat it for three times and report their average results. More details can be found in attached supplementary material.

\subsection{Ablation Study}
\begin{table}[t]
	\centering
	\scriptsize
	\begin{tabular}{l|c|c}
	\toprule
	Method & \multicolumn{2}{c}{Pretext Task} \\ \cline{2-3}
	& Rotation \cite{gidaris2018unsupervised} & Position \cite{doersch2015unsupervised} \\
	\hline
	ContraLoss & 85.5 & 82.3 \\
	{\footnotesize ContraLoss + Self-Recons.} & 87.9 & 84.5 \\
	{\footnotesize ContraLoss + TransAtt (Flip)} & 89.2 & 86.4 \\
	{\footnotesize ContraLoss + TransAtt} & 91.0 & 89.3 \\
	{\footnotesize ContraLoss + CrossMix} & 90.5 & 88.3 \\
	PCRL (All modules) & 93.2 & 91.6\\
	\bottomrule
	\end{tabular}
	\caption{Investigation of whether our method contains more information. \textbf{ContraLoss} stands for the contrastive loss. \textbf{Self-Recons.} represents self-reconstruction which is reconstructing the input images without any variations. \textbf{Acc.} stands for the classification accuracy. \textbf{TransAtt (Flip)} means that the indicator function in TransAtt only contains the flip operation.}
	\label{dis2}
\end{table}
\begin{table}[t]
%    \footnotesize
%    \small
    \centering
    \small
	\begin{tabular}{l|cc}
	\toprule
	\multirow{2}{*}{Method} & \multicolumn{2}{c}{\footnotesize Chest14$\rightarrow$Chest14} \\\cline{2-3}
	& 9:1 & 8:2 \\
	\hline
	ContraLoss & 71.7 & 74.8  \\
	\hline
	ContraLoss + Self-Recons. & 72.5 & 75.4  \\
	ContraLoss + RotNet & 72.3 & 75.2   \\
	ContraLoss + TransAtt (Flip) & 73.5 & 76.6 \\
	ContraLoss + TransAtt & 74.4 & 77.4 \\
	\hline
	ContraLoss + CrossMix ($\alpha$=0.5) & 73.3 & 76.1  \\
	ContraLoss + CrossMix ($\alpha$=1) & 73.7 & 76.6  \\
	\hline
 	PCRL (All modules) & 76.2 & 78.8  \\
	\bottomrule
	\end{tabular}
	\caption{Investigation of different module combinations. In Chest14$\rightarrow$Chest14, \textbf{9:1} demonstrates that we use 90\% data for self-supervised pretraining while the rest 10\% are used for finetuning. \textbf{RotNet} represents that we replace Self-Recons. with the task of rotation prediction.}
	\label{ab_chest14}
\end{table}
\begin{figure*}
	\centering
	\scriptsize
	\subfloat[2D tasks: pretraining using Chest14 or CheXpert]
	{
%	\begin{table}
	\scriptsize
% 	\small
	\begin{tabular}{l|c|c|c|c|c|c|c|c|c|c|c|c}
	\toprule
	\multirow{2}{*}{Method} & \multicolumn{5}{c|}{Chest14$\rightarrow$Chest14} & \multicolumn{7}{c}{CheXpert$\rightarrow$Chest14} \\ \cline{2-13}
	& 9.5:0.5 & 9:1 & 8:2 & 7:3 & 6:4 & 10\% & 20\% & 30\% & 40\% & 50\% & 60\% & 100\% \\
	\hline
	TS & 61.8 & 68.1 & 71.5 & 73.4 & 75.4 & 68.1 & 71.5 & 73.4 & 75.4 & 77.5 & 79.1  & 80.9\\
	\hline
	IN & 70.5 & 73.6 & 75.3 & 76.9 & 78.0 & 73.5 & 76.3 & 78.4 & 79.0 & 79.5 & 79.7 & 81.0 \\
	\hline
	MG & 66.4 & 70.0 & 73.9 & 76.1 & 77.3 & 70.1 & 73.9 & 75.5 & 76.5 & 77.6 & 79.3 & 80.8 \\
	\hline
	SG & 66.5 & 70.2 & 74.3 & 76.7 & 77.6 & 69.7 & 73.8 & 75.6 & 77.3 & 77.3 & 79.6 & 81.3 \\
	\hline
	C2L & 71.7 & 74.1 & 76.4 & 77.5 & 79.0 & 73.1 & 77.0 & 78.5 & 79.1 & 79.8 & 80.2 & 81.5 \\
	\hline
	PCRL & \textbf{74.1} & \textbf{76.2} & \textbf{78.8} & \textbf{79.0} & \textbf{79.9} & \textbf{75.8} & \textbf{77.6} & \textbf{79.8} & \textbf{80.8} & \textbf{81.2} & \textbf{81.7} & \textbf{83.1} \\
	\hline
	p-value & \footnotesize 5.2e-4 & \footnotesize 9.6e-4 & \footnotesize 2e-3 & \footnotesize 1.8e-3 & \footnotesize 2.3e-3 & \footnotesize 2.4e-3 & \footnotesize 8.1e-4 & \footnotesize 2.4e-3 & \footnotesize 3.5e-4 & \footnotesize 5.6e-4 & \footnotesize 3.6e-3 & \footnotesize 2.7e-3 \\
	\bottomrule
	\end{tabular}\label{2D}}		
%	\end{table}
	\\
	\subfloat[3D tasks: pretraining using LUNA]
	{
%	\begin{table}
	\scriptsize
% 	\small
	\begin{tabular}{l|c|c|c|c|c|c|c|c|c|c|c|c|c|c}
	\toprule
	\multirow{2}{*}{Method} & \multicolumn{4}{c|}{LUNA$\rightarrow$LUNA} & \multicolumn{5}{c|}{LUNA$\rightarrow$LiTS} & \multicolumn{5}{c}{LUNA$\rightarrow$BraTS} \\ \cline{2-15}
	& 9:1 & 8:2 & 7:3 & 6:4 & 10\% & 20\% & 30\% & 40\% & 100\% & 10\% & 20\% & 30\% & 40\% & 100\% \\
	\hline
	TS & 78.4 & 83.0 & 85.7 & 87.5 & 71.1 & 77.2 & 84.1 & 87.3 & 90.7 & 66.6 & 72.7 & 76.7 & 77.1 & 81.5 \\
	\hline
	MG & 80.2 & 85.0 & 87.5 & 90.3 & 73.3 & 79.5 & 84.3 & 87.9 & 91.3 & 69.6 & 75.5 & 79.6 & 80.4 & 82.4 \\
	\hline
	Cube++ & 81.4 & 85.2 & 87.9 & 90.0 & 74.2 & 79.3 & 84.5 & 88.2 & 91.8 & 69.0 & 74.9 & 79.3 & 79.7 & 82.2\\
	\hline
	SG & 79.3 & 84.5 & 87.9 & 90.5 & 73.8 & 79.3 & 85.5 & 88.2 & 91.4 & 70.3 & 75.6 & 79.1 & 80.8 & 82.3 \\
	\hline
	3D-CPC & 80.2 & 85.2 & 88.3 & 90.6 & 74.8 & 80.2 & 85.6 & 88.9 & 91.9 & 70.1 & 75.9 & 79.4 & 81.2 & 82.9 \\
	\hline
	PCRL & \textbf{84.4} & \textbf{87.5} & \textbf{89.8} & \textbf{92.2} & \textbf{77.3} & \textbf{83.5} & \textbf{87.8} & \textbf{90.1} & \textbf{93.7} & \textbf{71.6} & \textbf{77.6} & \textbf{81.1} & \textbf{83.3} & \textbf{85.0}\\
	\hline
	p-value & \footnotesize 7.5e-4 & \footnotesize 1.5e-3 & \footnotesize 2.1e-3 & \footnotesize 1.9e-3 & \footnotesize 2e-3 & \footnotesize 1.7e-3 & \footnotesize 9e-4 & \footnotesize 2.5e-3 & \footnotesize 2.4e-4 & \footnotesize 8.4e-4 & \footnotesize 3.5e-3 & \footnotesize 5.7e-3 & \footnotesize 2.5e-3 & \footnotesize 2.4e-3 \\
	\bottomrule
	\end{tabular}		
	\label{3D}
	}
	\caption{Comparison of different methods. In (a), we report the results of 2D tasks. In (b), the results of 3D tasks are displayed. The ratios in Chest14$\rightarrow$Chest14 and LUNA$\rightarrow$LUNA stand for the amount of unlabeled data (for pretraining) with respect to the amount of labeled data (for finetuning). In other tasks, the ratios represent the amount of data of the source dataset used for pretraining. For experiments of LiTS, we report the dice score of liver segmentation. For BraTS, we compute the mean dice of whole tumor, tumor core and enhancing tumor. We also report the p-values between the best and the second best results for each ratio to demonstrate the significance of PCRL.}
	\label{all}
\end{figure*}
\begin{figure*}[htp]
	\centering 
\includegraphics[width=2.0\columnwidth]{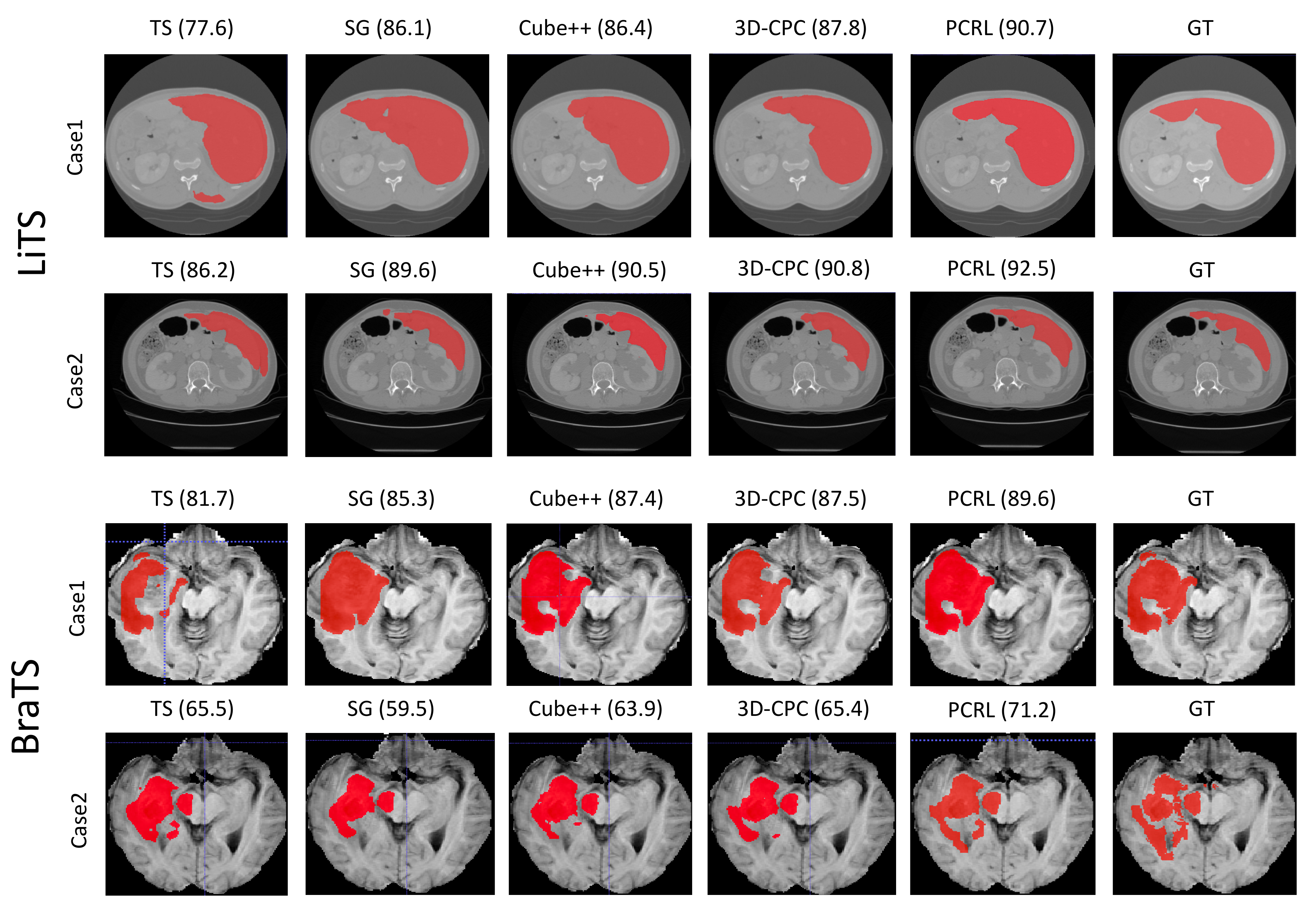}
	\caption{Visual analysis of segmentation results when finetuning on LiTS and BraTS. For each dataset, we provide 2 cases where we report the dice scores using different self-supervised pretraining methodologies. Specifically, in LiTS, the goal is to segment liver. In BraTS, we only display the results of WT. We ignore MG because SG is built on top of MG.}
	\label{visual}
\end{figure*}

In this section, we mainly investigate two problems: 1) whether the proposed method preserves more information than contrastive learning (Table \ref{dis2}) and 2) if the preserved information lead to the improved performance (Table \ref{ab_chest14}). In Table \ref{ab_chest14}, we make experiments on Chest14 to investigate the effectiveness of different module combinations, where we treat different ratios of the dataset as labeled data for supervised finetuning while the rest are used as unlabeled data for self-supervised pretraining.\\

\noindent \textbf{Preservational learning brings more information to representations.} In Table \ref{dis2}, we show that reconstructing diverse contexts does bring more information in learned representations. We introduce two pretext tasks: predicting the rotation degree \cite{gidaris2018unsupervised} and the relative position between images patches \cite{doersch2015unsupervised} to evaluate the amount of information in representations. In practice, we fixed the pretrained models as feature extractors and finetune the last fully connected layer for pretext tasks. Note that we directly utilize the pretrained models in CheXpert and use Chest14 to conduct pretext tasks. Specifically, when predicting the relative position between two image patches, we first divide the original input image to 14$\times$14 patches. Then, we extract adjacent image patches and formalize the position prediction problem as a 8-class classification problem (top, top left, top right, left, right, bottom left, bottom, bottom right). Similarly, when predicting the rotation degree, we manually rotate each input image by a specific degree and train the network to predict this degree, which can also be converted to a classification problem following \cite{gidaris2018unsupervised}. We display the classification results in Table \ref{dis2}. It is obvious that {ContraLoss + Self-Recons.} can already perform better than using {ContraLoss} only by preserving more information obtained from simply reconstructing the original input images. More importantly, the proposed {TransAtt} module outperforms {Self-Recons.} by only employing the flip operation. Together with the rotation transformations, {TransAtt} greatly surpasses {Self-Recons.} on both pretext tasks, which demonstrates that {TransAtt} is able to preserve much more information than reconstructing the original input images. Similar phenomena can also be observed when applying {CrossMix}. Finally, PCRL achieves much higher accuracy than the others, again verifying reconstructing diverse contexts do help preserve more information in learned features.\\

\noindent \textbf{Preserved information lead to better performance.}
We report the performance of different module combinations in Table \ref{ab_chest14}, where we can observe similar trends as those in Table \ref{dis2}. It is obvious that the results on pretext tasks are closely correlated with the performance on Chest14. In other words, given a method, we can rely on its performance on two pretext tasks to roughly predict its performance in Chest14. Considering the performance on two pretext tasks can reflect the amount of information in learned representations, we can easily draw a conclusion: \emph{reconstructing diverse contexts introduce more information which help improve the overall performance of algorithms.}

From Table \ref{ab_chest14}, we can easily find that adding a self-reconstruction branch only brings marginal improvements over the baseline model. Similar phenomena can also be observed when we replace Self-Recons. with RotNet \cite{gidaris2018unsupervised}. These results show that the contrastive loss already captures the information about simple pretext tasks without directly implementing these tasks. The fact that Self-Recons. performs better than RotNet shows that image reconstruction can preserve more information than RotNet. As for TransAtt, by comparing TransAtt (Flip) with TransAtt, we find that adding the rotation transformations can  obviously improve the overall performance. This is consistent with the results in Table \ref{dis2}, where TransAtt also performs better than TransAtt (Flip) on two pretext tasks. We also investigate the influence of the hyperparameter $\alpha$ in CrossMix. The observation is that by decreasing its value by half, the overall performance slightly drops. Equipped with TransAtt and CrossMix, PCRL can surpass the baseline model ContraLoss by approximate 4 points in different labeled ratios. Moreover, we find that the improvement is most significant at 10\%. This phenomenon implies that the reconstructing diverse contexts is more useful when the amount of labeled data is small.

\subsection{Comparison with State-of-the-Arts in 2D Tasks}
In this part, we evaluate the performance of various self-supervised pretraining approaches on 2 different 2D tasks: Chest14$\rightarrow$Chest14 and CheXpert$\rightarrow$Chest14. All results are displayed in Table \ref{2D}.

If we look at the results in Chest14$\rightarrow$Chest14, it is obvious that all pretraining methods (including IN) can boost the performance apparently when compared to TS. We can see that MG and SG achieve similar performance in different ratios. Such comparison is easy to explain as SG is built upon MG. However, both MG and SG still cannot surpass IN especially when the amount of labeled data is limited, which demonstrates that being pretrained on a large-scale natural image dataset can benefit medical image analysis a lot. As for C2L, we find that C2L is the only baseline method which is able to surpass IN in different ratios. When we compare PCRL with other baseline algorithms, it is easy to find that PCRL has the ability to outperform different baselines in various ratios significantly. Particularly, PCRL seems to have more advantages in small labeled ratios. The underlying reason may be that TransAtt and CrossMix may help to learn more diversified representations and alleviate the overfitting problem of training deep neural networks with limited supervision.

In CheXpert$\rightarrow$Chest14, we can see that MG and SG achieve comparable results with TS when the labeled ratio is equal or greater than 50\%, demonstrating purely pretext-based approaches may have unstable performance under varying labeled ratios. If we look at C2L, we can find that C2L consistently outperforms IN and other pretraining methods in almost all ratios. Somewhat surprisingly, we find that PCRL can still outperform C2L and IN by a significant margin even if the labeled ratio is 100\%. Such comparison further demonstrates the robustness of PCRL.

\begin{table}[t]
	\centering
	\scriptsize
	\begin{tabular}{l|c|c|c}
	\toprule
	Method & \#. Epoch & Cityscapes & COCO \\
	\hline
	SimCLR \cite{chen2019self} & 1000 & 75.6 & 39.6\\
	SwAV \cite{caron2018deep} & 400 & 76.0 & - \\
	MoCov2\cite{chen2020improved} & 800 & 76.3 & 40.5 \\
	PCRL & 800 & 77.3 & 41.3\\
	\bottomrule 	
	\end{tabular}
	\caption{Results in natural images. We transfer the self-supervised pretrained models on ImageNet-1k to downstream tasks, including segmentation (Cityscapes) and detection (COCO). On Cityscapes, we use ResNet-50 as backbone to build a FCN segmentation model where the evaluation metric is mIoU. On COCO, we use the ResNet-50-FPN model from Detectron2 \cite{wu2019detectron2} and the evaluation metric is mAP (0.5:0.05:0.95).}
	\label{natural}
\end{table}
\subsection{Comparison with State-of-the-Arts in 3D Tasks}
Besides 2D tasks, we also analyze the results of 3D self-supervised learning approaches in 3 different 3D tasks: LUNA$\rightarrow$LUNA, LUNA$\rightarrow$LiTS and LUNA$\rightarrow$BraTS, where all experimental results are shown in Table \ref{3D}.

In LUNA$\rightarrow$LUNA, it is interesting to find that the performance gaps between TS and self-supervised pretraining are smaller than those in Chest14. One explanation is that the nodule classification task is less sensitive to the amount of labeled data. Among MG, SG, Cube++ and 3D-CPC, 3D-CPC gives the best results in large labeled ratios while Cube++ performs better in small ones. Interestingly, as the labeled ratio increases, SG quickly catches up with MG and Cube++, showing its ability to utilize a large number of labeled images. Again, we can see that PCRL is able to outperform other baselines significantly in different ratios. Particularly, when the baseline approaches show similar results as the labeled ratio becomes larger, PCRL can still display impressive improvements over previous self-supervised pretraining approaches and outperforms TS substantially. In LUNA$\rightarrow$LiTS, Cube++ performs slightly better than MG and SG while 3D-CPC outperforms Cube++ in allmost all ratios. By comparison, PCRL has apparent advantages over other baselines especially when the labeled ratio is smaller or equal to 50\%. 

When we transfer knowledge from LUNA to BraTS, MG, SG and Cube++ display similar performance, all surpassing TS significantly in different labeled ratios. Due  to advantages of contrastive learning, 3D-CPC again outperforms other baselines. Meanwhile, PCRL once again surpasses previous baselines consistently and remarkably. We think that such significant improvements can be attributed to the incorporation of the reconstruction of diverse contexts.

\subsection{Visual Analysis}
 In Figure \ref{visual}, we provide comparative visual analysis results of segmentation tasks in LiTS and BraTS, where the samples are \emph{randomly} selected. We can obviously observe that PCRL handles the details much better than those of other baselines. For instance, in the first example of LiTS, PCRL delineates the corners accurately. In the second example of BraTS, PCRL can detect the isolated tumor regions while other methods cannot well handle these difficult cases.\\

\subsection{Comparison with State-of-the-Arts in Natural Image Segmentation and Detection Tasks}
To investigate the performance of PCRL in natural images, we conduct pretraining tasks on ImageNet-1k and transfer the pretrained models to downstream segmentation and detection tasks. The results are displayed in Table \ref{natural}. We can see that PCRL is able to outperform MoCov2 and other popular self-supervised learning methods substantially in both Cityscapes and COCO, which are two widely adopted datasets in segmentation and detection. The superior performance on Cityscapes and COCO again verify the advantages of incorporating diverse context reconstruction.

\section{Discussion and Conclusion}
We show that by reconstructing diverse contexts, the learned representations using the contrastive loss can be greatly improved in medical image analysis.  Our approach has shown positive results of self-supervised learning in a variety of medical tasks and datasets. There are some questions worth further discussing and verifying. For example, is preserving more information the only reason leading to the improvements over the contrastive loss? We hope the proposed PCRL can lay the foundations for real-world medical imaging tasks.

\section{Acknowledgements}
This work and the related project were funded in part by National Key Research and Development Program of China (No.2019YFC0118101), National Natural Science Foundation of China (No.81971616), Zhejiang Province Key Research \& Development Program (No.2020C03073), National Natural Science Foundation of China (No.61931024) and the Guangdong Provincial Key Laboratory of Big Data Computing, The Chinese University of Hong Kong, Shenzhen.

%%%%%%%%% BODY TEXT - ENTER YOUR RESPONSE BELOW
\section{Thorough Implementation Details (Supplementary Material)}
In each dataset, we build the training set, the validation set and the test set using 70\%, 10\% and 20\% of the whole dataset, respectively. Particularly, if a dataset is used for self-supervised pretraining and supervised finetuning in a semi-supervised manner, we construct the training set of pretraining by extracting a specific amount of data from the complete training set while the rest of the training set are used for finetuning. More details will be introduced in the following.
\begin{itemize}
	\item \textbf{Chest14 \cite{wang2017chestx}} is comprised of 112,120 X-ray images with 14 diseases from 30,805 unique patients. Each input image is resized to 224$\times$224 after random crop. In our experiments, Chest14 is used for both 2D pretraining and 2D finetuning. The batch size of pretraining is 256 and the batch size of finetuning is 512, where we use batch normalization for all experiments. The evaluation metric of finetuning is AUC.
	\item \textbf{CheXpert \cite{irvin2019chexpert}} is a large public dataset for chest radiograph interpretation, consisting of 224,316 chest radiographs of 65,240 patients. Each input image is resized to 224$\times$224 after random crop. Considering CheXpert contains uncertain labels, we mainly use CheXpert for 2D pretraining. The batch size of pretraining is 256 and we use batch normalization for all experiments.
%	\item \textbf{SegTHOR \cite{lambert2019segthor}} includes the segmentation of organs at risk in the thorax. We use 40 CT scans with available volumetric annotations for 2D supervised finetuning to study the problem of cross-modality transfer (from X-ray to CT). The evaluation metric is dice coefficient. For each input, we first apply center crop to remove blank areas. Then, we use random crop to generate 320$\times$320 image patches. The batch size of finetuning is 32 and we use batch normalization to stabilize the training process.
	\item \textbf{LUNA \cite{setio2017validation}} contains 888 CT scans with lung nodule annotations. Note that LUNA has different preprocessing strategies for different approaches which will be clarified in the following. We use LUNA for both 3D pretraining and 3D finetuning. The evaluation metric of finetuning is AUC. The range of the Hounsfield Units (HU) is [-1000,1000]. The batch size of pretraining and finetuning is 32 and we  use batch normalization in all experiments.
	\item \textbf{BraTS \cite{bakas2018identifying}} includes 351 MRI scans of brain tumor (in 2018 challenge). BraTS shares similar preprocessing strategies with those of LUNA. We use BraTS for 3D finetuning. The evaluation metric of finetuning is dice score. Since the batch size is 4 for all finetuning experiments, we use group normalization to replace batch normalization in the pretrained models.
	\item \textbf{LiTS \cite{bilic2019liver}} contains 131 available CT scans with their annotations for liver and tumor. LiTS is mainly used for 3D finetuning where we first localize the liver and expand the target volume by 30 slices in each axis. The segmentation target is liver only. The input size is 256$\times$256$\times$64 after random crop. Dice coefficient is employed as the evaluation metric while the range of HU is [-200,200]. The batch size is 4. Similar to BraTS, we also use group normalization to replace batch normalization in the pretrained models.
\end{itemize}
%\begin{table*}[htp]
%    \centering
%	\begin{tabular}{l|ccc|cc}
%	\toprule
%	\multirow{2}{*}{Vector}& \multicolumn{3}{c|}{Chest14$\rightarrow$Chest14} & \multicolumn{2}{c}{LUNA$\rightarrow$BraTS} \\ \cline{2-6}
%	& 9:1 & 8:2 & 7:3 & 10\% & 20\% \\
%	\hline
%	Flip + \{0,\ang{90},\ang{180},\ang{270}\} & 76.2 & 78.6 & 79.1 & 72.9 & 78.9 \\
%	\hline
%	Flip + \{0,\ang{90}\} & 75.8 & 78.2 & 78.6 & 72.4 & 78.2 \\
%	\hline
%	Flip + Random Degree & 76.4 & 78.5 & 78.8 & 73.1 & 78.8 \\
%	\hline
%	\{0,\ang{90},\ang{180},\ang{270}\} & 75.4 & 77.7 & 78.5 & 72.3 & 78.2 \\
%	\bottomrule
%	\end{tabular}
%	\caption{Investigation of variations of the indicator vector.}
%	\label{variations}
%\end{table*}

For network architectures, in Chest14$\rightarrow$Chest14 and CheXpert$\rightarrow$Chest14, we simply follow \cite{zhou2020comparing,sowrirajan2020moco} to conduct experiments based on ResNet-18. 
%Similarly, in CheXpert$\rightarrow$SegTHOR, we build an U-Net like architecture based on ResNet-18. 
In 3D tasks, we directly use the ordinary 3D U-Net as backbone. We will demonstrate the implementation details of each approach in the following.
\begin{itemize}
	\item \textbf{MG \cite{zhou2020models}}. In LUNA and BraTS datasets, we apply random crop to produce each input tensor whose size is 64$\times$64$\times$32 (corresponding to x, y and z axes). The augmentation strategies on all datasets include patch shuffling, non-linear transformation, inpainting, outpainting and random flip.
	\item \textbf{SG \cite{haghighi2020learning}}. Similar to MG, we apply random crop to generate 64$\times$64$\times$32 patches in LUNA and BraTS datasets. For the number of nearest neighbors, we set it to 200/1000 in 3D and 2D tasks, respectively. Note that SG follows similar augmentation strategies used in MG.
	\item \textbf{Cube++ \cite{tao2020revisiting}}. Similar to MG and SG, the input size is 64$\times$64$\times$32 in LUNA and BraTS datasets. Then, we divide each volume into 32 subvolumes. The size of each subvolume is 16$\times$16$\times$16. We then apply random rotation to each subvolume and employs a generative neural network to reconstruct the original input volume.
	\item \textbf{C2L \cite{zhou2020comparing}}. Give a 2D input image, we apply random crop, flip, rotation (10 degree) and grayscale to generate random patches. For each patch, we resize it to 224$\times$224 and apply inpainting to it.
	\item \textbf{3D-CPC \cite{cciccek20163d}}. We follow the instructions in \cite{cciccek20163d} to implement a 3D-CPC model. 
	\item \textbf{PCRL}. The augmentation strategies include random crop, random flip, random rotation, inpainting, outpainting and gaussian blur. For 2D images, we simply apply these operations to generate two input patches. Specially, for each 3D input, we randomly crop 16 pairs of volumetric inputs. The two volumes in each pair have the same size which is randomly sampled from \{64$\times$64$\times$32, 96$\times$96$\times$48, 96$\times$96$\times$96, 32$\times$32$\times$16, 112$\times$112$\times$56\}. Note that we require the Interaction over Union (IoU) within each pair to be larger than 0.2. Finally, we resample each volume to 64$\times$64$\times$32\footnote{We only use LUNA for self-supervised pretraining in 3D volumes.}. Note that random rotation is only applied to xy-plane in 3D volume where the maximum degree is \ang{10}. The length of queue which stores past representations is 16384 in PCRL. Note that the reason why we do not use online random crop in 3D images is that the computational cost of such operation (including crop and resize) is somewhat unacceptable in our GPU device. For both 2D and 3D tasks, we take 4 NVIDIA TITAN V GPUs to train PCRL models. The pretraining tasks on Chest14 and LUNA take 1 day and a half day, respectively.
\end{itemize}

{\small
\bibliographystyle{ieee_fullname}
\bibliography{egbib}
}

\end{document}